\documentclass[11pt,a4paper]{article}
\pdfoutput=1
\usepackage{coling2020}
\usepackage{times}
\usepackage{latexsym}
\usepackage{booktabs}
\usepackage{multirow}
\usepackage{url}
\usepackage{float}
\usepackage{threeparttable}

\usepackage{soul}

\usepackage{CJKutf8}
\usepackage{xcolor}
\usepackage{graphicx}

\usepackage{balance}
\usepackage{stfloats}
\usepackage{longtable}
\usepackage{xtab,afterpage}

\usepackage{microtype}
\usepackage{fancyhdr}

\colingfinalcopy

\title{CLUE: A Chinese Language Understanding Evaluation Benchmark}

\author{
    \begin{tabular}{c} 
        Liang Xu, Hai Hu, Xuanwei Zhang, Lu Li, Chenjie Cao, Yudong Li, Yechen Xu, Kai Sun,  \\
        Dian Yu, Cong Yu, Yin Tian, Qianqian Dong, Weitang Liu, Bo Shi, Yiming Cui, Junyi Li,\\
        Jun Zeng, Rongzhao Wang, Weijian Xie, Yanting Li, Yina Patterson, Zuoyu Tian,\\
        Yiwen Zhang, He Zhou, Shaoweihua Liu, Zhe Zhao, Qipeng Zhao, Cong Yue, \\
        Xinrui Zhang, Zhengliang Yang, Kyle Richardson and Zhenzhong Lan\thanks{$^\ast$ Corresponding author. E-mail: lanzhenzhong@westlake.edu.cn} \\
    \end{tabular} 
    \\
    CLUE team \\
    \texttt{CLUE@CLUEbenchmarks.com}
}

\begin{document}
\let\citep\cite
\let\citet\newcite

\maketitle
\begin{abstract}



The advent of natural language understanding (NLU) benchmarks for English, such as GLUE and SuperGLUE allows new NLU models to be evaluated across a diverse set of tasks. These comprehensive benchmarks have facilitated a broad range of research and applications in natural language processing (NLP). The problem, however, is that most such benchmarks are limited to English, which has made it difficult to replicate many of the successes in English NLU for other languages. To help remedy this issue, we introduce the first large-scale Chinese Language Understanding Evaluation (CLUE) benchmark. CLUE is an open-ended, community-driven project that brings together 9 tasks spanning several well-established single-sentence/sentence-pair classification tasks, as well as machine reading comprehension, all on original Chinese text. To establish results on these tasks, we report scores using an exhaustive set of current state-of-the-art pre-trained Chinese models (9 in total). We also introduce a number of supplementary datasets and additional tools to help facilitate further progress on Chinese NLU. Our benchmark is released at \textcolor{blue}{\url{https://www.CLUEbenchmarks.com}}

\end{abstract}
\section{Introduction}


Full-network pre-training methods such as BERT \citep{bert-2019} and their improved versions \citep{XLNet,liu2019roberta,ALBERT} have led to significant performance boosts across many natural language understanding (NLU) tasks. One key driving force behind such improvements and rapid iterations of models is the general use of evaluation benchmarks. These benchmarks use a single metric to evaluate the performance of models across a wide range of tasks. However, existing language evaluation benchmarks are mostly in English, e.g.,~GLUE \citep{wang2018glue} and SuperGLUE \citep{wang2019superglue}. To the best of our knowledge, there is no general language understanding evaluation benchmark for Chinese, whose speakers account for one-fourth of the world's population.  
Also, Chinese is linguistically very different from English and other Indo-European languages, which necessitates an evaluation benchmark specifically designed for Chinese. Without such a benchmark, it would be difficult for researchers in the field to check how good their Chinese language understanding models are.



To address this problem and facilitate studies in Chinese language, we introduce a comprehensive \textbf{C}hinese \textbf{L}anguage \textbf{U}nderstanding \textbf{E}valuation (CLUE) benchmark that contains a collection of nine different natural language understanding tasks (two of which are created by us), including semantic similarity, natural language inference, short text classification, long text classification with large number of classes, and different types of machine reading comprehension tasks.
To better understand the challenges posed by these tasks, we evaluate them using several popular pre-trained language understanding models for Chinese. Overall, we find that these tasks display different levels of difficulty,  manifest in different accuracies across models, as well as the comparison between human and machine performance. 




The size and quality of unlabeled corpora play an essential role in language model pre-training \citep{bert-2019,XLNet,liu2019roberta,ALBERT}. There are already popular pre-training corpora such as Wikipedia and the Toronto Book Corpus~\citep{zhu2015aligning} in English. However, we are not aware of any large-scale open-source pre-training dataset in Chinese. Also Chinese models are mainly trained on different and relatively small corpora. Therefore, it is difficult to improve model performance and compare them across model architectures. This difficulty motivates us to construct and release a standard CLUE pre-training dataset: a corpus with over $214$ GB raw text and roughly $76$ billion Chinese words.
We also introduce a diagnostic dataset hand-crafted by linguists. 
Similar to GLUE, this dataset is designed to highlight linguistic and common knowledge and logical operators that we expect models to handle well.

Overall, we present in this paper:
(1) A Chinese natural language understanding benchmark that covers a variety of sentence classification and machine reading comprehension tasks, at different levels of difficulty, in different sizes and forms.  
(2) A large-scale raw corpus for general-purpose pre-training in Chinese so that the comparisons across different model architectures are as meaningful as possible.  
(3) A diagnostic evaluation dataset developed by linguists containing multiple linguistic phenomena, some of which are unique to Chinese.
(4) A user-friendly toolkit, as well as an online leaderboard with an auto-evaluation system, supporting all our evaluation tasks and models, with which researchers can reproduce experimental results and compare the performance of different submitted models easily. 

\section{Related Work}
\label{sec:related_work}

It has been a common practice to evaluate language representations on different intrinsic and downstream NLP tasks. 
For example, \citet{mikolov2013distributed} measure word embeddings through a semantic analogy task and a syntactic analogy task. \citet{pennington2014glove} further expands the testing set to include other word similarity and named entity recognition tasks. 
Similar evaluation procedures are also used for sentence representations \citep{kiros2015skip}. 
However, as different researchers use different evaluation pipelines on different datasets, results reported in the papers are not always fully comparable, especially in the case where the datasets are small, where a minor change in evaluation can lead to big differences in outcomes. 

SentEval \citep{conneau2018senteval} addresses the above problem by introducing a standard evaluation pipeline using a set of popular sentence embedding evaluation datasets. GLUE \citep{wang2018glue} and SuperGLUE \citep{wang2019superglue} further improve SentEval by providing benchmarks for natural language understanding tasks, ensuring that results from different models are consistent and comparable. They introduce a set of more difficult datasets and a model-agnostic evaluation pipeline. Along with other reading comprehension tasks like SQuAD \citep{rajpurkar-etal-2016-squad} and RACE \citep{lai2017race-arxiv}, GLUE and SuperGLUE have become standard testing benchmarks for pre-training methods such as BERT \citep{bert-2019} and ALBERT \citep{ALBERT}. 

We believe a similar problem exists in Chinese language understanding evaluation. Although more and more Chinese linguistic tasks \citep{liu2018lcqmc,cui2018span} have been proposed, there is still a need for a standard evaluation pipeline and an evaluation benchmark with a set of diverse and difficult language understanding tasks.  

\section{CLUE Overview}
\label{sec:overview}

CLUE consists of 1) nine language understanding tasks in Chinese, 2) a large-scale raw dataset for pre-training and a small hand-crafted diagnostic dataset for linguistic analysis, and 3) a ranking system, a leaderboard and a toolkit.


\subsection{Task Selection}
For this benchmark, we selected nine different tasks, to ensure that the benchmark tests different aspects of pre-trained models. 
To ensure the quality and coverage of the language understanding tasks, we select tasks using the following criteria: 
\paragraph{Diversity}
The tasks in CLUE should vary in terms of the task, the size of the text, the type of understanding required, the number of training examples. 
\paragraph{Well-defined and easy-to-process}
We select tasks that are well-defined, and we pre-process them for our users so that they can focus on modeling.
\paragraph{Moderate difficulty: challenging but solvable} 
To be included in CLUE, a task should not be too simple or already solved so as to encourage researchers to design better models (e.g., multiple-choice machine reading comprehension task).

\paragraph{Representative and useful}
Our tasks should be representative of common language understanding tasks, easily applicable to real-world situations (e.g., classification task with many labels, or semantic similarity task). 


\paragraph{Tailor to Chinese-specific characteristics}
Ideally, tasks should measure the ability of models to handle Chinese-specific linguistic phenomena (e.g., four-character idioms).  


Although Chinese is not a low-resource language, it is still non-trivial to find and collect NLU tasks in Chinese, given a lack of diverse publicly available NLP datasets relative to English. Therefore apart from scrutinizing existing literature, we also sent out a call-for-tasks to the Chinese NLP community from which we received proposals or suggestions for several new datasets.\footnote{We only accepted some of them because other tasks were either not well-defined, or are normally not counted as NLU tasks (e.g., named-entity recognition).}   
In addition, to help overcome the lack of publicly-available NLU-oriented sentence-/sentence-pair classification tasks for Chinese, we created two new tasks for our benchmark (CLUEWSC2020 and CSL, see section \ref{sec:tasks} for details).
Based on the above standards, we gathered a total of nine tasks in the end, seven of them selected from our collected datasets plus two newly created by us. These tasks cover a broad range of text genres, linguistic phenomena and task-formats. 

 

\subsection{Large-scale Pre-Training Dataset} 
We collect data from the internet and preprocess them to make a large pre-training dataset for Chinese language processing researchers. In the end, a total of $214$ GB raw corpus with around $76$ billion Chinese words are collected in our pre-training corpus (see Section \ref{sec:pretraining_dateset} for details). 
 
\subsection{Diagnostic Dataset}
In order to measure how well models are doing on specific language understanding phenomena, we handcraft a diagnostic dataset that contains nine linguistic and logic phenomena (details in Section \ref{sec:diagnostics}).


\subsection{Leaderboard}
We also provide a leaderboard for users to submit their own results on CLUE. The evaluation system will give final scores for each task when users submit their predicted results. To encourage reproducibility, we mark the score of a model as ``certified'' if it is open-source, and we can reproduce the results.



\subsection{Toolkit}
To make it easier for using the CLUE benchmark, we also offer a toolkit named PyCLUE implemented in TensorFlow \citep{abadi2016tensorflow}. PyCLUE supports mainstream pre-training models and a wide range of target tasks. Different from existing pre-training model toolkits \citep{wolf2019transformers,zhao2019uer}, PyCLUE is designed with a goal of quick model performance validations on the CLUE benchmark. 
\section{Tasks}
\label{sec:tasks}

\begin{table*}[t]
\centering \small
\begin{tabular}{lrrrp{4cm}ll}
    \toprule
        \textbf{Corpus} & \textbf{$|$Train$|$} & \textbf{$|$Dev$|$} & \textbf{$|$Test$|$} & \textbf{Task} & \textbf{Metric} & \textbf{Source} \\
    \midrule
        \multicolumn{7}{c}{\bf Single-Sentence Tasks}\\
    \midrule
        TNEWS & 53.3k & 10k & 10k &  short text classification & acc. & news title and keywords \\
        IFLYTEK & 12.1k & 2.6k & 2.6k &long text classification & acc. & app descriptions\\
        CLUEWSC2020 & 1,244 & 304 & 290 & coreference resolution & acc. & Chinese fiction books \\
    \midrule
        \multicolumn{7}{c}{\bf Sentence Pair Tasks}\\
    \midrule
        AFQMC & 34.3k & 4.3k & 3.9k & semantic similarity & acc. & online customer service \\
        CSL & 20k & 3k & 3k & keyword recognition & acc. & academic (CNKI) \\
        OCNLI & 50k & 3k & 3k & natural language inference & acc. & 5 genres  \\
    \midrule    
    \multicolumn{7}{c}{\bf Machine Reading Comprehension Tasks}\\
    \midrule
        CMRC 2018 & 10k & 3.4k & 4.9k & answer span extraction & EM.
        &  Wikipedia \\
        ChID & 577k & 23k & 23k    & multiple-choice, idiom  & acc. & novel, essay, and news \\
        C$^3$ & 11.9k  & 3.8k & 3.9k & multiple-choice, free-form & acc. & mixed-genre \\
        
    \bottomrule
\end{tabular}
\caption{Task descriptions and statistics. TNEWS has $15$ classes; IFLYTEK has $119$ classes; OCNLI has $3$ classes, other classification tasks are binary classification.}
\label{tab:tasks description}
\end{table*}

CLUE has nine Chinese NLU tasks, covering single sentence classification, sentence pair classification, and machine reading comprehension. Descriptions of these tasks are shown in Table \ref{tab:tasks description}, and examples of these are shown in Table \ref{tab:examples} in the Appendix.

\subsection{Single Sentence Tasks}
\paragraph{TNEWS}
TouTiao Text Classification for News Titles\footnote{\urlstyle{same}\url{https://github.com/fatecbf/toutiao-text-classfication-dataset/}} consists of Chinese news published by TouTiao before May 2018, 
with a total of 73,360 titles.
Each title is labeled with one of $15$ news categories (finance, technology, sports, etc.)
and the task is to predict which category the title belongs to. To make the dataset more discriminative, we use cross-validation to filter out some of the easy examples (see Section D Dataset Filtering in the Appendix for details). We then randomly shuffle and split the whole dataset into a training set, development set and test set.

\paragraph{IFLYTEK}
IFLYTEK \citep{2019iflytek} contains
17,332 app descriptions. The task is to assign each description into one of $119$ categories, such as food, car rental, education, etc. A data filtering technique similar to the one used for the TNEWS dataset has been applied.
  
\paragraph{CLUEWSC2020}
The Chinese Winograd Schema Challenge dataset
is an anaphora/coreference resolution task where the model is asked to decide whether a pronoun and a noun (phrase) in a sentence co-refer (binary classification), built following similar datasets in English (e.g., \citet{wsc} and \citet{wang2019superglue}). Sentences in the dataset are hand-picked from 36 contemporary literary works in Chinese. Their anaphora relations are then hand-annotated by linguists, amounting to 1,838 questions in total. 
  
\subsection{Sentence Pair Tasks}
Tasks in this section ask a model to predict relations between sentence pairs, or abstract-keyword pairs. 

\paragraph{AFQMC}
The Ant Financial Question Matching Corpus\footnote{\urlstyle{same}\url{https://dc.cloud.alipay.com/index\#/topic/intro?id=3}} comes from Ant Technology Exploration Conference (ATEC) Developer competition. It is a binary classification task that aims to predict whether two sentences are semantically similar.

\paragraph{CSL}
Chinese Scientific Literature dataset contains Chinese paper abstracts and their keywords from core journals of China, covering multiple fields of natural sciences and social sciences. We generate fake keywords through tf-idf and mix them with real keywords. Given an abstract and some keywords, the task is to tell whether the keywords are all original keywords of a paper. It mainly evaluates the ability of models to judge whether keywords can summarize the document. 

\paragraph{OCNLI} Original Chinese Natural Language Inference (OCNLI, \citet{ocnli}) is collected closely following procedures of MNLI \citep{mnli}. OCNLI is composed of 56k inference pairs from five genres: news, government, fiction, TV transcripts and Telephone transcripts, where the premises are collected from Chinese sources, and universities students in language majors are hired to write the hypotheses. The annotator agreement is on par with MNLI. We believe the non-translation nature of OCNLI makes it more suitable than XNLI \citep{conneau2018xnli} as an NLU task specific for Chinese. 

\subsection{Machine Reading Comprehension}
\paragraph{CMRC 2018}
CMRC 2018 \citep{cui2018span} is a span-extraction based dataset for Chinese machine reading comprehension. This dataset contains about 19,071 human-annotated questions from  Wikipedia paragraphs. In CMRC 2018, all samples are composed of contexts, questions, and related answers. Furthermore, the answers are the text spans in contexts.
 

 \paragraph{ChID}
 ChID \citep{zheng2019chid} is a large-scale Chinese IDiom cloze test dataset, which contains about 498,611 passages with 623,377 blanks covered from news, novels, and essays. The candidate pool contains 3,848 Chinese idioms. For each blank in the passage, there are ten candidate idioms with one golden option, several similar idioms, and others are randomly chosen from the dictionary. 

\paragraph{C\textsuperscript{3}}
C\textsuperscript{3}~\citep{sun2019probing} is the first free-form multiple-choice machine reading comprehension dataset for Chinese. Given a document, either a dialogue or a more formally written mixed-genre text, and a free-form question that is not limited to a single question type (e.g., yes/no questions), the task is to select the correct answer option from all (2 to 4) options associated with the corresponding question.  We employ all of the 19,577 general domain problems for 13,369 documents and follow the original data splitting. These problems are collected from language exams carefully designed by educational experts for evaluating the reading comprehension ability of language learners, similar to its English counterparts RACE~\citep{lai2017race-arxiv} and DREAM~\citep{sun2019dream}.

\section{Pre-Training Dataset}
\label{sec:pretraining_dateset}


Large-scale language data is the prerequisite for model pre-training. 
Corpora of various sizes have been compiled and utilized in English, e.g., the Wikipedia Corpus, the BooksCorpus \citep{zhu2015aligning}, and more recent C4 corpus \citep{2019t5}.

For Chinese, however, existing public pre-training datasets are much smaller than the English datasets. For example, the Wikipedia dataset in Chinese only contains around $1.1$ GB raw text.
We thus collect a large-scale clean crawled Chinese corpus to fill this gap.

A total of $214$ GB raw corpus with around 76 billion words are collected, consisting of three different corpora: CLUECorpus2020-small, CLUECorpus2020, and CLUEOSCAR. Three models in this paper are pre-trained on the combined CLUE pre-training corpus (two ALBERT models and RoBERTa-large).

\paragraph{CLUECorpus2020-small} It contains 14 GB of Chinese text, with the following genres:


\begin{itemize}
\setlength\itemsep{.01em}

\item \textbf{News}~~This sub-corpus  is crawled from the We Media  (self-media) platform, with a total of $3$ billion Chinese words from $2.5$ million news articles of roughly $63$K sources. 

\item \textbf{WebText}~~With $4.1$ million questions and answers, the WebText sub-corpus is crawled from  Chinese Reddit-like websites such as Wukong QA, Zhihu, Sogou Wenwen, etc. Only answers with three or more upvotes are included to ensure the quality of the text.

\item \textbf{Wikipedia}~~This sub-corpus is gathered from the Chinese contents on Wikipedia (Chinese Wikipedia), containing around $1.1$ GB raw texts with $0.4$ billion Chinese words on a wide range of topics. 

\item \textbf{Comments}~~These comments are collected from E-commerce websites including Dianping.com and Amazon.com by SophonPlus\footnote{\urlstyle{same}\url{https://github.com/SophonPlus/ChineseNlpCorpus/}}. This subset has approximately $2.3$ GB of raw texts with $0.8$ billion Chinese words.
\end{itemize}

\paragraph{CLUECorpus2020} 

It contains 100 GB Chinese raw corpus, which is   retrieved   from Common Crawl. It is a well-defined  dataset  that  can  be  used  directly for pre-training without requiring additional pre-processing. CLUECorpus2020 contains around 29K separate files with each file following the pre-training format for the training set.

\paragraph{CLUEOSCAR\footnote{\urlstyle{same}\url{https://dumps.wikimedia.org/zhwiki/latest/}}}

OSCAR is a huge multilingual corpus obtained by language classification and filtering of the Common Crawl corpus. It contains 250 GB Chinese raw corpus. We do further filtering and finally get 100 GB Chinese corpus.



\section{Experiments}
\label{sec:experiments}
\begin{table*}[t]
\footnotesize
\centering \fontsize{8.4}{10.1}\selectfont \setlength{\tabcolsep}{0.5em}
\scalebox{0.95}{
\begin{tabular}{l|c|ccc|ccc|ccc}

    \toprule

    & & \multicolumn{3}{|c|}{\textbf{Single Sentence}} & \multicolumn{3}{|c|}{\textbf{Sentence Pair}} & \multicolumn{3}{|c}{\textbf{MRC}} \\
    \midrule
    
\textbf{Model} & \textbf{Avg} & \textbf{TNEWS} & \textbf{IFLYTEK} & \textbf{CLUEWSC2020} & \textbf{AFQMC} &  \textbf{CSL} &  \textbf{OCNLI}  & \textbf{CMRC}  & \textbf{ChID} & \textbf{C$^3$}\\
    \midrule


    BERT-base & 69.20 & 56.58 & 60.29 & 63.45 & 73.70 & 80.36 & 72.20 & 69.72 & 82.04 & 64.50\\ 
    BERT-wwm-ext-base & 70.27 & 56.84 & 59.43 & 62.41 & 74.07 & 80.63 & 74.42 & 73.23  & 82.90 & 68.50\\ 
    
    ALBERT-tiny & 56.01 &  53.35 & 48.71 & 63.38 & 69.92 & 74.56 & 65.12 & 53.68 & 43.53 & 31.86\\ 
    
    ALBERT-xxlarge & 72.49 & \underline{59.46} & 62.89 & 61.54 & 75.60 & \underline{83.63}  & 77.70 & 75.15 & 83.15 & \underline{73.28} \\
    
    ERNIE-base & 69.72  &  58.33 & 58.96 & 63.44 & 73.83 & 79.10 & 74.11 & 73.32 & 82.28 & 64.10\\ 

    XLNet-mid & 68.58 &  56.24 & 57.85 & 61.04 & 70.50 & 81.26 & 72.63 & 66.51  & 83.47 & 67.68\\ 

    RoBERTa-large & 71.01  &  57.86 & 62.55 & 62.44 & 74.02 & 81.36 & 76.82 & 76.11  & 84.50 & 63.44\\ 
    
    RoBERTa-wwm-ext-base & 71.17 &  56.94 & 60.31 & 72.07 & 74.04 & 81.00 & 74.72 & 73.89  & 83.62 & 63.90\\ 
    
        
    RoBERTa-wwm-ext-large & \underline{74.90} &  58.61 & \underline{62.98} & \underline{81.38} & \underline{76.55} & 82.13 & \underline{78.20} & \underline{76.58}  & \underline{85.37} & 72.32 \\ 
    
    \midrule
    Human & \textbf{85.09} &  \textbf{71.00} & \textbf{66.00} & \textbf{98.00} & \textbf{81.0} & \textbf{84.0} & \textbf{90.30}   & \textbf{92.40} & \textbf{87.10} & \textbf{96.00} \\
    \bottomrule
\end{tabular}
}
\caption{Performance of baseline models on CLUE benchmark.
\textit{Avg} is the average of all tasks. \textbf{Bold} text denotes the best result in each column. \underline{Underline} indicates the best result for the models. We report EM for CMRC 2018 and  accuracy for all other tasks. 
} 
\label{tab:benchmark}
\end{table*}

 
 \paragraph{Baselines}
 
Our baseline models are built on different pre-trained transformers \cite{vaswani2017attention}, on which an additional output layer is added for fine-tune on CLUE tasks. 
For single-sentence tasks, we encode the sentence and then pass the pooled output to a classifier. For sentence-pair tasks, we encode sentence pairs with a separator and then pass the pooled output to a classifier. As for the  extraction-style and multi-choice style for machine reading comprehension tasks, we use two fully connected layers after the pooled output to predict the start and end position of the answer for the former. For the latter, we encode multiple candidate-context pairs to a shared classifier and get corresponding scores.

All the models are implemented in both TensorFlow \citep{abadi2016tensorflow} and PyTorch \citep{paszke2019pytorch}. 
 
 \paragraph{Models} 
 
 We evaluate CLUE on the following public available pre-trained models: 
 
 
 
\begin{itemize}
\setlength\itemsep{0.01em}

\item BERT-base, we use the base model (12 layer, hidden size 768) published by \cite{bert-2019}, which was pre-trained the on Chinese Wikipedia dump of about 0.4 billion tokens.
\item BERT-wwm-ext-base, a model with the same configuration of BERT-base except it uses whole word masking and is trained on additional 5 billion tokens \citep{chinese-bert-wwm}.

\item ALBERT-tiny/xxlarge, ALBERT \citep{ALBERT} is a recent language representation model. We use: 1) a tiny version\footnote{\urlstyle{same}\url{https://github.com/brightmart/albert_zh}} with only 4 layers and a hidden size of 312, and 2) an xxlarge version\footnote{\urlstyle{same}\url{https://github.com/google-research/albert}} with 12 layers and a hidden size of 4096. Both are trained on the CLUE pre-training corpus.


\item ERNIE-base \citep{ernie1.0} extends BERT-base with additional training data and leverages knowledge from Knowledge Graphs. 

\item XLNet-mid\footnote{\urlstyle{same}\url{https://github.com/ymcui/Chinese-PreTrained-XLNet}},  a model with 24 layers and a hidden size of 768, with sentencepiece tokenzier and other techniques from \citet{XLNet}. 


\item RoBERTa-large uses a 24 layer RoBERTa  \citep{liu2019roberta} with a hidden size of 1024, trained with the CLUE pre-training corpus. 

\item RoBERTa-wwm-ext-base \citep{chinese-bert-wwm} uses a 12 layer Transformer \citep{vaswani2017attention} with a hidden size of 768,  it uses whole word masking and is trained on the same dataset as BERT-base-wwm except following the training procedure of \citet{liu2019roberta}. 

\item RoBERTa-wwm-ext-large \citep{chinese-bert-wwm} has a network structure of RoBERTa-large and training procedure of RoBERTa-wwm-ext-base.



\end{itemize}

We believe these models are representative of most of the current transformer architectures. In particular, ALBERT-xxlarge and RoBERTa-wwm-ext-large are the largest models in Chinese at the time of writing, and are expected to give us an estimate of the upperbound of model performance. We include ALBERT-tiny to examine empirically how big the performance reduction is when switched to a much smaller model, which presents another estimate for scenarios with limited computing resources.
A summary of the hyper-parameters of these models can be found in Table \ref{tab:app:para:pretrain} in the Appendix.

\paragraph{Fine-tuning} 
We fine-tune the pre-trained models separately for each task. Hyper-parameters are chosen based on the performance of each model on the development set. We also use early stopping to select the best checkpoint. Each model is fine-tuned three times and we choose the model with the best performance on the development set to report test results. 


\subsection{Human Performance}

    
    

OCNLI, CMRC 2018, ChID and C$^3$  have provided human performance \citep{ocnli,sun2019probing,cui2018span,zheng2019chid}. 
For those tasks without human performance in CLUE, we ask human annotators to label 100 randomly chosen items from the test set and compute the annotators' majority vote against the gold label.

\begin{table*}[!t]
\small
\centering 
    \centering
    \begin{tabular}{llrrrrr}
     \toprule
        & & \textbf{TNEWS}  & \textbf{AFQMC} & \textbf{CSL} &  \textbf{IFLYTEK} &  \textbf{CLUEWSC2020}\\
    \midrule
       \multirow{5}{*}{\shortstack{Trained \\ annotation}} 
       & annotator 1 & 57.0 & 83.0 & 93.0   & 54.0 & 94.0 \\
       & annotator 2 & 66.0 & 81.0 & 80.0   & 80.0 & 97.0 \\
       & annotator 3 & 73.0 & 76.0 & 67.0   & 50.0 & 95.0 \\
    \cmidrule{2-7}
       & avg & 65.3 & 80.0 & 80.0 & 61.3 & 95.3  \\
       & majority & \textbf{71.0} & \textbf{81.0} & \textbf{84.0} & \textbf{66.0} & \textbf{98.0}\\
    \midrule
        & best model & 58.61 & 76.5 & 82.13 & 62.98 & 81.38 \\
    \bottomrule 
    \end{tabular}
    \caption{Two-stage human performance scores and the best accuracy of models comparison. ``avg'' denotes the mean score from the three annotators. ``majority'' shows the performance if we take the majority vote from the labels given by the annotators. \textbf{Bold} text denotes the best result among human and model performance.}
    \label{tab:human:perf:two-stage}
\end{table*}



We follow procedures in SuperGLUE \citep{wang2019superglue} to train the annotators before asking them to work on the test data. 
Specifically, each annotator is first asked to annotate 30 to 50 pieces of data from the development set, and then compare their labels with the gold ones. They are then encouraged to discuss their mistakes and questions with other annotators until they are confident about the task. Then they annotate 100 pieces of test data, which is used to compute our final human performance, shown in 
Table \ref{tab:human:perf:two-stage} and the last row of Table \ref{tab:benchmark}. 
As we can see, most of the tasks are relatively easy for humans with a score in the 80s and 90s, except for TNEWS and IFLYTEK, both of which have many classes, potentially making it harder for humans.
We will discuss human performance in light of the models' performance in the next section.


\subsection{Benchmark Results}
We report the results of our baseline models on the CLUE benchmark in Table \ref{tab:benchmark}.

\paragraph{Analysis of Model Performance}

The first thing we notice is that the results are better when: 1) the model is larger, or 2) the model is trained with more pre-training data, or 3) whole word masking is used. Specifically, RoBERTa-wwm-ext-large and ALBERT-xxlarge are the two best performing models, showing advantages over other models particularly for machine reading tasks such as C$^3$. 

Next, we want to highlight the results from ALBERT-tiny, which has only about 1/20 of the parameters in BERT-base model. Our results suggest that for single-sentence or sentence-pair tasks, the performance drop compared with BERT-base can range from almost 0 (for CLUEWSC2020) to roughly 12 percentage points (IFLYTEK). 
However, for tasks involving more global understanding, small models have more serious limitations,
as illustrated by ALBERT-tiny's low accuracy in all three machine reading tasks, with a performance drop of up to 40 percentage compared with BERT-base (ChID). 

Finally, XLNet-mid, a model based on a common unsupervised tokenizer in English called SentencePiece \citep{kudo2018sentencepiece}, performs poorly in token level Chinese tasks like span-extraction based MRC (CMRC 2018). This highlights the need for our Chinese-specific benchmark which provides empirical results as to whether successful techniques in English can be readily applied or transferred to a very different language such as Chinese, where no word boundaries are present in running texts.

\paragraph{Analysis of Tasks}
It seems that what is easy for human may not be so for machine. For instance, humans are very accurate in multiple-choice reading comprehension (C$^3$), whereas machines struggle in it (ALBERT-tiny has a very low accuracy of about 32\%, probably due to the small size of the model). The situation is similar for CLUEWSC2020, where the best score of models is far behind human performance (about 17 percentage points). Note that in SuperGLUE, RoBERTa did very well on the English WSC (89\% against 100\% for humans), whereas in our case, the performance of variants of RoBERTa is still much lower than the average human performance, though it is better than other models.
 

On the other hand, tasks such as CSL and ChID seem to be of equal difficulty for humans and machines, with accuracies in the 80's for both. For humans, the keyword judgment task (CSL) is hard because the fake keywords all come from the abstract of the journal article, which has many technical terms. Annotators are unlikely to perform well when working with unfamiliar jargon.

Surprisingly, the hardest dataset for both humans and machines is a single sentence task: TNEWS. One possible reason is that news titles can potentially fall under multiple categories (e.g., finance and technology) at the same time, while there is only one gold label in TNEWS. 

The best result from machines remains far below human performance, with roughly 11 points lower than human performance on average. This leaves much room for further improvement of models and methods, which we hope will drive the Chinese NLP community forward. 
\section{Diagnostic Dataset for CLUE}
\label{sec:diagnostics}
\paragraph{Dataset Creation}

\begin{CJK*}{UTF8}{gbsn}

\begin{table*}[t]
\small
\centering
\scalebox{.84}{
\begin{tabular}{p{1.9cm}|l|p{4.2cm}p{4.2cm}|l|lll|rrr}
\toprule
 &  &  &  &  & \multicolumn{3}{c|}{\textbf{Predictions}} & \multicolumn{3}{c}{\textbf{Accuracy}} \\
 & \# & \textbf{Premise} & \textbf{Hypothesis} & \rotatebox[origin=c]{90}{~gold}  & BE & RO & XL  & BE & RO & XL \\\hline
Anaphora & 48 & 马丽和她的母亲李琴一起住在这里。\newline Ma Li and her mother Li Qin live here together. & 马丽是李琴的母亲。\newline Ma Li is Li Qin's mother. & C & E & E & E & 47.9 & 58.3 & 47.9 \\\hline
Argument structure & 50 & 小白看见小红在打游戏。\newline Xiao Bai saw Xiao Hong playing video games. & 小红在打太极拳。\newline Xiao Hong is doing Tai Chi. & C & C & C & C & 60.0 & 60.0 & 54.0 \\\hline
Common sense & 50 & 小明没有工作。\newline Xiaoming doesn't have a job. & 小明没有住房。\newline Xiaoming doesn't have a place to live. & N & N & N & C & 44.0 & 58.0 & 48.0 \\\hline
Comparative & 50 & 这筐桔子比那筐多。\newline This basket has more oranges than that one. & 这筐桔子比那筐多了不少。\newline This basket has much more oranges than that one. & N & E & E & E & 36.0 & 56.0 & 46.0 \\\hline
Double negation & 24 & 你别不把小病小痛当一回事。\newline Don't take minor illness as nothing. & 你应该重视小病小痛。\newline You should pay attention to minor illness. & E & E & E & E & 54.2 & 62.5 & 62.5 \\\hline
Lexical semantics & 100 & 小红很难过。\newline Xiaohong is sad. & 小红很难看。\newline Xiaohong is ugly. & N & E & N & E & 62.0 & 70.0 & 64.0 \\\hline
Monotonicity & 60 & 有些学生喜欢在公共澡堂里唱歌。\newline Some students like to sing in the shower room. & 有些女生喜欢在公共澡堂里唱歌。\newline Some female students like to sing in the shower room. & N & N & N & N & 41.7 & 43.3 & 43.3 \\\hline
Negation & 78 & 女生宿舍，男生勿入。\newline Girls dormitory, no entering for boys. & 女生宿舍只能女生进出。\newline Only girls can go in and out of the girls dormitory. & E & E & C & C & 62.8 & 64.1 & 60.3 \\\hline
Time of event & 54 & 记者去年采访企业家了。\newline The reporter interviewed the entrepreneur last year. & 记者经常采访企业家。\newline The reporter interviews the entrepreneur very often. & N & N & N & N & 61.1 & 74.1 & 59.3 \\\hline
Total &  &  &  &  &  &  &  & 53.5 & 61.5 & 54.7 \\
\bottomrule
\end{tabular}
}
\caption{The CLUE diagnostics: Example test items in 9 linguistic categories, with their gold labels and model predictions, as well as model accuracy.  E = entailment, N = neutral, C = contradiction. BE = BERT-base, RO = RoBERTa-wwm-ext-large, XL = XLNet-mid.
\label{tab:diagnostics}}
\end{table*}

In order to examine whether the trained models can master linguistically important and meaningful phenomena, we follow GLUE \citep{wang2018glue} to provide a diagnostic dataset, setting up as a natural language inference task and predicting whether a  hypothesis is \textit{entailed} by, \textit{contradicts} to or is \textit{neutral} to a given premise. Crucially, we did not translate the English diagnostics into Chinese, as the items in their dataset may be specific to English language or American/Western culture. 
Instead, we have several Chinese linguists hand-crafting 514 sentence pairs in idiomatic Chinese from scratch. These pairs cover 9 linguistic phenomena and are manually labeled by the same group of linguists. We ensured that the labels are balanced (majority baseline is 35.1\%). Examples are shown in Table \ref{tab:diagnostics}. 
Some of the categories directly address the unique linguistic properties of Chinese. For instance, items in the ``Time of event'' category test models on their ability to handle aspect markers such as 着~(imperfective marker), 了~(perfective marker), 过~(experiential marker), which convey information about the time of event, whether it is happening now or has already happened in the past. We believe that for a model to make robust inferences, it needs to understand such unique Chinese phenomena, and also has other important linguistic abilities, such as handling anaphora resolution \citep{webster2018mind} and monotonicity reasoning \citep{yanaka2019can,probe2020}. 
\end{CJK*}

\paragraph{Evaluation and Error Analysis}

\begin{CJK*}{UTF8}{gbsn}
We evaluate three representative models on the diagnostic dataset: BERT-base, XLNet-mid,  RoBERTa-wwm-ext-large. Each model is fine-tuned on OCNLI, and then tested on our diagnostic dataset. As illustrated in Table \ref{tab:diagnostics}, the highest accuracy is only about 61\%, which indicates that models have a hard time solving these linguistically challenging problems. We believe that both models and inference datasets suggest room for improvement.

A breakdown of results is presented in the last few columns of Table \ref{tab:diagnostics}. Monotonicity is the hardest, similar to GLUE diagnostics \citep{wang2018glue}. It seems that BERT also has a hard time dealing with comparatives.  
An interesting case is the example of lexical semantics in Table \ref{tab:diagnostics}, where the two two-character words ``sad'' (难过~\textit{hard-pass}) and ``ugly'' (难看~\textit{hard-look}) in Chinese have the same first character (难~\textit{hard}). 
Thus the premise and hypothesis only differ in the last character, which two out of three models have decided to ignore. One possible explanation is that these models in Chinese are also using the simple lexical overlap heuristic, as illustrated in \citet{hans} for English.
\end{CJK*}

\section{Conclusions and Future Work}
\label{sec:discussions}


In this paper, we present a Chinese Language Understanding Evaluation (CLUE) benchmark, which consists of 9 natural language understanding tasks and a linguistically motivated diagnostic dataset, along with an online leaderboard for model evaluation. In addition, we release a large clean crawled raw text corpus that can be directly used for pre-training Chinese models. To the best of our knowledge, CLUE is the first comprehensive language understanding benchmark developed for Chinese. We evaluate several latest language representation models on CLUE and analyze their results.  An analysis is conducted on the diagnostic dataset created by Chinese linguists, which illustrates the limited ability of state-of-the-art models to handle some Chinese linguistic phenomena.

In contrast to the English benchmarks such as GLUE and SuperGLUE, where model performance is already at human performance, we can see that Chinese NLU still has considerable room for improvement (i.e., models are $\sim$10\% below our estimates of human performance), meaning that we expect that our benchmark will facilitate building better models in the short-term. Once models have reached human performance, however, we 
believe that extending our benchmark to newer tasks, or newer forms of evaluation (e.g., taking into account performance as a function of model size as in \cite{lightmodel2020}), could be a step forward. In this sense, we view CLUE, which is an entirely community-driven project, to be open-ended in that our current set of tasks serve as a first step in more comprehensively evaluating Chinese NLU.

\section{Acknowledgement}
\label{sec:acknowledge}
The authors would like to thank everyone who has contributed their datasets to CLUE. We are also grateful to the annotators and engineers who have spent much of their time and effort helping with the creation of the CLUE benchmark. Special thanks to the following companies and organizations: OneConnect Financial Technology Co., Ltd, OpenBayes Co., Ltd, AI-Indeed.com, Alibaba Cloud Computing, Joint Laboratory of HIT and iFLYTEK Research (HFL). Research supported with Cloud TPUs from Google's TensorFlow Research Cloud (TFRC). 


\bibliography{coling2020}

\begin{thebibliography}{}

\bibitem[\protect\citename{Abadi \bgroup et al.\egroup
  }2016]{abadi2016tensorflow}
Mart{\'\i}n Abadi, Ashish Agarwal, Paul Barham, Eugene Brevdo, Zhifeng Chen,
  Craig Citro, Greg~S Corrado, Andy Davis, Jeffrey Dean, Matthieu Devin, et~al.
\newblock 2016.
\newblock Tensorflow: Large-scale machine learning on heterogeneous distributed
  systems.
\newblock {\em arXiv preprint arXiv:1603.04467}.

\bibitem[\protect\citename{Conneau and Kiela}2018]{conneau2018senteval}
Alexis Conneau and Douwe Kiela.
\newblock 2018.
\newblock {SentEval}: An evaluation toolkit for universal sentence
  representations.
\newblock In {\em Proceedings of the Eleventh International Conference on
  Language Resources and Evaluation (LREC 2018)}.

\bibitem[\protect\citename{Conneau \bgroup et al.\egroup
  }2018]{conneau2018xnli}
Alexis Conneau, Ruty Rinott, Guillaume Lample, Adina Williams, Samuel~R.
  Bowman, Holger Schwenk, and Veselin Stoyanov.
\newblock 2018.
\newblock {XNLI}: Evaluating cross-lingual sentence representations.
\newblock In {\em Proceedings of the 2018 Conference on Empirical Methods in
  Natural Language Processing}. Association for Computational Linguistics.

\bibitem[\protect\citename{Cui \bgroup et al.\egroup }2019]{cui2018span}
Yiming Cui, Ting Liu, Wanxiang Che, Li~Xiao, Zhipeng Chen, Wentao Ma, Shijin
  Wang, and Guoping Hu.
\newblock 2019.
\newblock A span-extraction dataset for {C}hinese machine reading
  comprehension.
\newblock In {\em Proceedings of the 2019 Conference on Empirical Methods in
  Natural Language Processing and the 9th International Joint Conference on
  Natural Language Processing (EMNLP-IJCNLP)}, pages 5886--5891, Hong Kong,
  China, November. Association for Computational Linguistics.

\bibitem[\protect\citename{Cui \bgroup et al.\egroup }2020]{chinese-bert-wwm}
Yiming Cui, Wanxiang Che, Ting Liu, Bing Qin, Shijin Wang, and Guoping Hu.
\newblock 2020.
\newblock Revisiting pre-trained models for chinese natural language
  processing.
\newblock In {\em Findings of EMNLP}. Association for Computational
  Linguistics.

\bibitem[\protect\citename{Devlin \bgroup et al.\egroup }2019]{bert-2019}
Jacob Devlin, Ming-Wei Chang, Kenton Lee, and Kristina Toutanova.
\newblock 2019.
\newblock {BERT}: Pre-training of deep bidirectional transformers for language
  understanding.
\newblock In {\em Proceedings of the 2019 Conference of the North {A}merican
  Chapter of the Association for Computational Linguistics: Human Language
  Technologies, Volume 1 (Long and Short Papers)}, pages 4171--4186,
  Minneapolis, Minnesota, June. Association for Computational Linguistics.

\bibitem[\protect\citename{Hu \bgroup et al.\egroup }2020]{ocnli}
Hai Hu, Kyle Richardson, Xu~Liang, Li~Lu, Sandra K\"{u}bler, and Larry Moss.
\newblock 2020.
\newblock {OCNLI}: Original {Chinese} natural language inference.
\newblock In {\em Findings of Empirical Methods for Natural Language Processing
  (Findings of EMNLP)}.

\bibitem[\protect\citename{IFLYTEK~CO.}2019]{2019iflytek}
LTD. IFLYTEK~CO.
\newblock 2019.
\newblock Iflytek: a multiple categories chinese text classifier.
\newblock {\em competition official website,
  http://challenge.xfyun.cn/2019/gamelist}.

\bibitem[\protect\citename{Kiros \bgroup et al.\egroup }2015]{kiros2015skip}
Ryan Kiros, Yukun Zhu, Russ~R Salakhutdinov, Richard Zemel, Raquel Urtasun,
  Antonio Torralba, and Sanja Fidler.
\newblock 2015.
\newblock Skip-thought vectors.
\newblock In {\em Advances in neural information processing systems}, pages
  3294--3302.

\bibitem[\protect\citename{Kudo and Richardson}2018]{kudo2018sentencepiece}
Taku Kudo and John Richardson.
\newblock 2018.
\newblock {SentencePiece}: A simple and language independent subword tokenizer
  and detokenizer for neural text processing.
\newblock In {\em Proceedings of the 2018 Conference on Empirical Methods in
  Natural Language Processing: System Demonstrations}, pages 66--71.

\bibitem[\protect\citename{Lai \bgroup et al.\egroup }2017]{lai2017race-arxiv}
Guokun Lai, Qizhe Xie, Hanxiao Liu, Yiming Yang, and Eduard Hovy.
\newblock 2017.
\newblock {RACE: Large-scale ReAding Comprehension Dataset From Examinations}.
\newblock In {\em Proceedings of the 2017 Conference on Empirical Methods in
  Natural Language Processing}, pages 785--794.

\bibitem[\protect\citename{Lan \bgroup et al.\egroup }2019]{ALBERT}
Zhenzhong Lan, Mingda Chen, Sebastian Goodman, Kevin Gimpel, Piyush Sharma, and
  Radu Soricut.
\newblock 2019.
\newblock {ALBERT: A Lite BERT for Self-supervised Learning of Language
  Representations}.
\newblock In {\em International Conference on Learning Representations}.

\bibitem[\protect\citename{Levesque \bgroup et al.\egroup }2012]{wsc}
Hector Levesque, Ernest Davis, and Leora Morgenstern.
\newblock 2012.
\newblock The winograd schema challenge.
\newblock In {\em Thirteenth International Conference on the Principles of
  Knowledge Representation and Reasoning}.

\bibitem[\protect\citename{Li \bgroup et al.\egroup }2020]{lightmodel2020}
Junyi Li, Hai Hu, Xuanwei Zhang, Minglei Li, Lu~Li, and Liang Xu.
\newblock 2020.
\newblock Light pre-trained {Chinese} language model for nlp tasks.
\newblock In Xiaodan Zhu, Min Zhang, Yu~Hong, and Ruifang He, editors, {\em
  Natural Language Processing and Chinese Computing}, pages 567--578. Springer
  International Publishing.

\bibitem[\protect\citename{Liu \bgroup et al.\egroup }2018]{liu2018lcqmc}
Xin Liu, Qingcai Chen, Chong Deng, Huajun Zeng, Jing Chen, Dongfang Li, and
  Buzhou Tang.
\newblock 2018.
\newblock Lcqmc: A large-scale chinese question matching corpus.
\newblock In {\em Proceedings of the 27th International Conference on
  Computational Linguistics}, pages 1952--1962.

\bibitem[\protect\citename{Liu \bgroup et al.\egroup }2019]{liu2019roberta}
Yinhan Liu, Myle Ott, Naman Goyal, Jingfei Du, Mandar Joshi, Danqi Chen, Omer
  Levy, Mike Lewis, Luke Zettlemoyer, and Veselin Stoyanov.
\newblock 2019.
\newblock {RoBERTa}: A robustly optimized bert pretraining approach.
\newblock {\em arXiv preprint arXiv:1907.11692}.

\bibitem[\protect\citename{McCoy \bgroup et al.\egroup }2019]{hans}
Tom McCoy, Ellie Pavlick, and Tal Linzen.
\newblock 2019.
\newblock Right for the wrong reasons: Diagnosing syntactic heuristics in
  natural language inference.
\newblock In {\em Proceedings of the 57th Annual Meeting of the Association for
  Computational Linguistics}, pages 3428--3448, Florence, Italy, July.
  Association for Computational Linguistics.

\bibitem[\protect\citename{Mikolov \bgroup et al.\egroup
  }2013]{mikolov2013distributed}
Tomas Mikolov, Ilya Sutskever, Kai Chen, Greg~S Corrado, and Jeff Dean.
\newblock 2013.
\newblock Distributed representations of words and phrases and their
  compositionality.
\newblock In {\em Advances in neural information processing systems}, pages
  3111--3119.

\bibitem[\protect\citename{Paszke \bgroup et al.\egroup
  }2019]{paszke2019pytorch}
Adam Paszke, Sam Gross, Francisco Massa, Adam Lerer, James Bradbury, Gregory
  Chanan, Trevor Killeen, Zeming Lin, Natalia Gimelshein, Luca Antiga, et~al.
\newblock 2019.
\newblock Pytorch: An imperative style, high-performance deep learning library.
\newblock In {\em Advances in Neural Information Processing Systems}, pages
  8024--8035.

\bibitem[\protect\citename{Pennington \bgroup et al.\egroup
  }2014]{pennington2014glove}
Jeffrey Pennington, Richard Socher, and Christopher Manning.
\newblock 2014.
\newblock Glove: Global vectors for word representation.
\newblock In {\em Proceedings of the 2014 conference on empirical methods in
  natural language processing (EMNLP)}, pages 1532--1543.

\bibitem[\protect\citename{Raffel \bgroup et al.\egroup }2020]{2019t5}
Colin Raffel, Noam Shazeer, Adam Roberts, Katherine Lee, Sharan Narang, Michael
  Matena, Yanqi Zhou, Wei Li, and Peter~J. Liu.
\newblock 2020.
\newblock Exploring the limits of transfer learning with a unified text-to-text
  transformer.
\newblock {\em Machine Learning Research,}, pages 1--67.

\bibitem[\protect\citename{Rajpurkar \bgroup et al.\egroup
  }2016]{rajpurkar-etal-2016-squad}
Pranav Rajpurkar, Jian Zhang, Konstantin Lopyrev, and Percy Liang.
\newblock 2016.
\newblock {SQ}u{AD}: 100,000+ questions for machine comprehension of text.
\newblock In {\em Proceedings of the 2016 Conference on Empirical Methods in
  Natural Language Processing}, pages 2383--2392, Austin, Texas, November.
  Association for Computational Linguistics.

\bibitem[\protect\citename{Richardson \bgroup et al.\egroup }2020]{probe2020}
Kyle Richardson, Hai Hu, Lawrence~S Moss, and Ashish Sabharwal.
\newblock 2020.
\newblock Probing natural language inference models through semantic fragments.
\newblock In {\em Proceedings of AAAI}.

\bibitem[\protect\citename{Sun \bgroup et al.\egroup }2019a]{sun2019dream}
Kai Sun, Dian Yu, Jianshu Chen, Dong Yu, Yejin Choi, and Claire Cardie.
\newblock 2019a.
\newblock Dream: A challenge data set and models for dialogue-based reading
  comprehension.
\newblock {\em Transactions of the Association for Computational Linguistics},
  7:217--231.

\bibitem[\protect\citename{Sun \bgroup et al.\egroup }2019b]{sun2019probing}
Kai Sun, Dian Yu, Dong Yu, and Claire Cardie.
\newblock 2019b.
\newblock Probing prior knowledge needed in challenging chinese machine reading
  comprehension.
\newblock {\em CoRR}, cs.CL/1904.09679v2.

\bibitem[\protect\citename{Sun \bgroup et al.\egroup }2019c]{ernie1.0}
Yu~Sun, Shuohuan Wang, Yukun Li, Shikun Feng, Xuyi Chen, Han Zhang, Xin Tian,
  Danxiang Zhu, Hao Tian, and Hua Wu.
\newblock 2019c.
\newblock {ERNIE}: Enhanced representation through knowledge integration.
\newblock {\em arXiv preprint arXiv:1904.09223}.

\bibitem[\protect\citename{Vaswani \bgroup et al.\egroup
  }2017]{vaswani2017attention}
Ashish Vaswani, Noam Shazeer, Niki Parmar, Jakob Uszkoreit, Llion Jones,
  Aidan~N Gomez, {\L}ukasz Kaiser, and Illia Polosukhin.
\newblock 2017.
\newblock Attention is all you need.
\newblock In {\em Advances in neural information processing systems}, pages
  5998--6008.

\bibitem[\protect\citename{Wang \bgroup et al.\egroup }2018]{wang2018glue}
Alex Wang, Amanpreet Singh, Julian Michael, Felix Hill, Omer Levy, and Samuel
  Bowman.
\newblock 2018.
\newblock {GLUE}: A multi-task benchmark and analysis platform for natural
  language understanding.
\newblock In {\em Proceedings of the 2018 {EMNLP} Workshop {B}lackbox{NLP}:
  Analyzing and Interpreting Neural Networks for {NLP}}, pages 353--355,
  Brussels, Belgium, November. Association for Computational Linguistics.

\bibitem[\protect\citename{Wang \bgroup et al.\egroup }2019]{wang2019superglue}
Alex Wang, Yada Pruksachatkun, Nikita Nangia, Amanpreet Singh, Julian Michael,
  Felix Hill, Omer Levy, and Samuel~R Bowman.
\newblock 2019.
\newblock Superglue: A stickier benchmark for general-purpose language
  understanding systems.
\newblock {\em Neural Information Processing Systems}, pages 3266--3280.

\bibitem[\protect\citename{Webster \bgroup et al.\egroup
  }2018]{webster2018mind}
Kellie Webster, Marta Recasens, Vera Axelrod, and Jason Baldridge.
\newblock 2018.
\newblock Mind the gap: A balanced corpus of gendered ambiguous pronouns.
\newblock {\em Transactions of the Association for Computational Linguistics},
  6:605--617.

\bibitem[\protect\citename{Williams \bgroup et al.\egroup }2018]{mnli}
Adina Williams, Nikita Nangia, and Samuel Bowman.
\newblock 2018.
\newblock A broad-coverage challenge corpus for sentence understanding through
  inference.
\newblock In {\em Proceedings of the 2018 Conference of the North American
  Chapter of the Association for Computational Linguistics: Human Language
  Technologies, Volume 1 (Long Papers)}, pages 1112--1122.

\bibitem[\protect\citename{Wolf \bgroup et al.\egroup
  }2019]{wolf2019transformers}
Thomas Wolf, Lysandre Debut, Victor Sanh, Julien Chaumond, Clement Delangue,
  Anthony Moi, Pierric Cistac, Tim Rault, R{\'e}mi Louf, Morgan Funtowicz,
  et~al.
\newblock 2019.
\newblock Transformers: State-of-the-art natural language processing.
\newblock {\em arXiv preprint arXiv:1910.03771}.

\bibitem[\protect\citename{Yanaka \bgroup et al.\egroup }2019]{yanaka2019can}
Hitomi Yanaka, Koji Mineshima, Daisuke Bekki, Kentaro Inui, Satoshi Sekine,
  Lasha Abzianidze, and Johan Bos.
\newblock 2019.
\newblock Can {N}eural {N}etworks {U}nderstand {M}onotonicity {R}easoning?
\newblock In {\em ACL Workshop BlackboxNLP}.

\bibitem[\protect\citename{Yang \bgroup et al.\egroup }2019]{XLNet}
Zhilin Yang, Zihang Dai, Yiming Yang, Jaime Carbonell, Russ~R Salakhutdinov,
  and Quoc~V Le.
\newblock 2019.
\newblock {XLNet}: Generalized autoregressive pretraining for language
  understanding.
\newblock In {\em Advances in neural information processing systems}, pages
  5753--5763.

\bibitem[\protect\citename{Zhao \bgroup et al.\egroup }2019]{zhao2019uer}
Zhe Zhao, Hui Chen, Jinbin Zhang, Wayne~Xin Zhao, Tao Liu, Wei Lu, Xi~Chen,
  Haotang Deng, Qi~Ju, and Xiaoyong Du.
\newblock 2019.
\newblock {UER: An Open-Source Toolkit for Pre-training Models}.
\newblock In {\em Proceedings of the 2019 Conference on Empirical Methods in
  Natural Language Processing and the 9th International Joint Conference on
  Natural Language Processing (EMNLP-IJCNLP): System Demonstrations}, pages
  241--246.

\bibitem[\protect\citename{Zheng \bgroup et al.\egroup }2019]{zheng2019chid}
Chujie Zheng, Minlie Huang, and Aixin Sun.
\newblock 2019.
\newblock {ChID: A Large-scale Chinese IDiom Dataset for Cloze Test}.
\newblock In {\em Proceedings of the 57th Annual Meeting of the Association for
  Computational Linguistics}, pages 778--787.

\bibitem[\protect\citename{Zhu \bgroup et al.\egroup }2015]{zhu2015aligning}
Yukun Zhu, Ryan Kiros, Rich Zemel, Ruslan Salakhutdinov, Raquel Urtasun,
  Antonio Torralba, and Sanja Fidler.
\newblock 2015.
\newblock Aligning books and movies: Towards story-like visual explanations by
  watching movies and reading books.
\newblock In {\em Proceedings of the IEEE international conference on computer
  vision}, pages 19--27.

\end{thebibliography}
\bibliographystyle{coling}

\clearpage
\appendix
\label{sec:appendix}

\section{Dataset Samples}
We have compiled examples of each data set for your reference in Table \ref{tab:examples}. Some of them are intercepted because the sentences are too long. For the complete data sets, you can refer to related papers. We will also release the download link of those datasets in the final version of the paper.

\section{Additional Parameters}
\label{sec:appendix-parameters}

\subsection{Hyperparameters for pre-training} 


Although we did not train most of the models by ourselves, we list the hyperparameter for pre-training in Table \ref{tab:app:para:pretrain} for reference purpose.

\subsection{Hyperparameters for fine-tuning}

Hyperparameters for fine-tuning in our experiments are listed in Table  \ref{tab:app:para:fine-tuning}.  

\section{Additional Baseline Details}
\label{sec:appendix-experiments-details}
\paragraph{CSL}
In generating negative samples for CSL, we only replace one of the real keywords with a fake one. When fine-tuning on CSL task, we found that some of the larger models can only converge at very small learning rates, for example, 5e-6.



\paragraph{IFLYTEK}
There are 126 categories in the original IFLYTEK dataset. However, some of them have few examples. We excluded those classes that have less than 10 examples so that we can apply the cross-validation filtering techniques as described in Section \ref{sec:appendix-filter-method}. During the experiments, we also found when fine-tuning Albert-tiny requires a larger number of epochs to converge compare to other models. Also, sentences in IFLYTEK dataset are relatively long compared to other sentence classification tasks. However, most of the useful information is located at the beginning of the sentences. We, therefore, choose a max length of 128.


\section{Dataset Filtering}
\label{sec:appendix-filter-method}
In order to increase the model differentiation and the difficulty of the dataset, we use four-fold cross-validation to filter iFLYTEK and TNEWS dataset. We divide the data sets in to four and use three of them to fine-tune ALBERT-tiny. After that, the fine-tuned model is used to select and filter those easy examples in the remaining set.

\begin{CJK}{UTF8}{gbsn}
\begin{table*}[!bp]
\small
\centering
\scalebox{1}{
\begin{tabular}{p{0.005\textwidth}p{0.93\textwidth}}
 \toprule
 \parbox[t]{1mm}{\multirow{2}{*}{\rotatebox[origin=c]{90}{{\textbf{TNEWS}}}}} &
\textbf{sentence:} 
如果我的世界下架了，你会玩迷你世界吗？
\\ &
\textbf{sentence (en):}
\textit{
If Minecraft is gone, will you play miniworld?
}
\\ & \textbf{label:} \texttt{116}(news\_game) \\

\midrule
\parbox[t]{1mm}{\multirow{2}{*}{\rotatebox[origin=c]{90}{{\textbf{iFLYTEK}}}}} &
\textbf{sentence:} 
《钢铁英雄》是一款角色扮演类游戏。游戏拥有 ...... 带领他们逃出去。修复部分小错误，提升整体稳定性。
\\ &
\textbf{sentence (en):}
\textit{
"Heroes of Steel" is a role-playing game. The game has ...... all four heroes are imprisoned and you will lead them out. repair part small
Errors to improve overall stability.
}
\\ & \textbf{label:} \texttt{22}(Strategy) \\

\midrule
\parbox[t]{1mm}{\multirow{2}{*}{\rotatebox[origin=c]{90}{{\textbf{CLUEWSC}}}}} &
\textbf{text:} 
这时候放在床上枕头旁边的\underline{手机}响了，我感到奇怪，因为欠费已被停机两个月，现在\underline{它}突然响了。
\\ &
\textbf{text (en):}
\textit{
At this moment, the \underline{cellphone} on the bed next to the pillow rang. I feel this is quite strange because the cellphone plan was terminated two months ago since I did not pay the bill. Now \underline{it} was ringing all of a sudden. 
}
\\ & \textbf{label:} \texttt{true} \\

\midrule
\parbox[t]{1mm}{\multirow{2}{*}{\rotatebox[origin=c]{90}{{\textbf{AFQMC}}}}} &
\textbf{sentence1:}
本月花呗还不上怎么办
\ 
\textbf{sentence2:}
花呗超时怎么办
\\&
\textbf{sentence1 (en):}
\textit{What to do if Ant Credit Pay is not available yet this month
} \ 
\textbf{sentence2 (en):}
\textit{How to deal with Ant Credit Pay overtime
}
\\ & \textbf{label:} \texttt{0}(different) \\

\midrule
\parbox[t]{1mm}{\multirow{2}{*}{\rotatebox[origin=c]{90}{{\textbf{CSL}}}}} &
\textbf{abst:}
不同阶段电子数据的操作都会留下表现各异的轨迹.从操作系统、计算机应用系统 ...... 分析审计电子数据轨迹在计算机系统中表现形式,可以为审计人员提供有效的审计方法
\\&
\textbf{keyword:}
[``计算机审计'', ``数据轨迹'', ``日志文件'']
\\&
\textbf{abst (en):}
\textit{
The operation of electronic data in different stages will leave different traces. From operating system, computer application system ...... provide effective audit methods for auditors by analyzing the expression of audit electronic data trace in computer system.
}\\&
\textbf{keyword (en):}
\textit{[``computer audit'', ``data trace'', ``log file'']
}
\\ & \textbf{label:} \texttt{0}(false) \\

\midrule
\parbox[t]{1mm}{\multirow{2}{*}{\rotatebox[origin=c]{90}{{\textbf{OCNLI}}}}} &
\textbf{premise:}
但是不光是中国,日本,整个东亚文化都有这个特点就是被权力影响很深
\ 
\textbf{hypothesis:}
有超过两个东亚国家有这个特点
\\&
\textbf{premise (en):}
\textit{But not only China and Japan, the entire East Asian
culture has this feature, that is it is deeply influenced by
the power.
} \ 
\textbf{hypothesis (en):}
\textit{More than two East Asian countries have this feature.
}
\\ & \textbf{label:} \texttt{entailment} \\

\midrule
\parbox[t]{1mm}{\multirow{2}{*}{\rotatebox[origin=c]{90}{{\textbf{CMRC 2018}}}}} &
\textbf{context:}
萤火虫工作室是一家总部设在英国伦敦和康涅狄格州坎顿...... 目前，他们正在开发PC和Xbox360上的次时代游戏。 
\\&
\textbf{question:} 萤火虫工作室的总部设在哪里？ \ 
\textbf{answer:} 英国伦敦和康涅狄格州坎顿
\\&
\textbf{context (en):}
\textit{Firefly Studios is a video game developer based in London, UK and Canton, Connecticut, with a quality department in Aberdeen, Scotland ...... Currently, they are developing next-generation games on PC and Xbox 360.}  
\\&
\textbf{question (en):} \textit{Where is Firefly Studios headquartered?} \
\textbf{answer (en):}
\textit{London, UK and Canton, Connecticut} 
\\

\midrule
\parbox[t]{1mm}{\multirow{2}{*}{\rotatebox[origin=c]{90}{{\textbf{ChID}}}}} &
\textbf{content:}
中国青年报：篮协改革联赛切莫\underline{\#idiom\#}......
\\&
\textbf{candidates:} [``急功近利'', ``画蛇添足'', ``\underline{本末倒置}''(answer)]
\\&
\textbf{content (en):}
\textit{China Youth Daily: Chinese Basketball Association should not \underline{\#idiom\#} when reforming the league ......}
\\&
\textbf{candidates (en): }
\textit{[``seeking instant benefit'', ``to overdo it'', ``\underline{take the branch for the root}''(answer)]
}
\\

\midrule
\parbox[t]{1mm}{\multirow{2}{*}{\rotatebox[origin=c]{90}{{\textbf{C\textsuperscript{3}}}}}} &
\textbf{document:}
男：我们坐在第七排，应该能看清楚字幕吧? 女：肯定可以，对了，我们得把手机设成振动。\\&
\textbf{question:} 他们最可能在哪儿?
\\&
\textbf{candidates:} [``图书馆'', ``体育馆'',``\underline{电影院}''(answer),``火车站'']
\\&
\textbf{document (en):}
\textit{Man: Our seats are in the seventh row. We should be able to see the subtitles clearly, right? Woman: Absolutely. By the way, we should set the phone to vibrate.
}\\&
\textbf{question (en)}: \textit{Where does the conversation most probably take place?}
\\&
\textbf{candidates (en):}
\textit{[``In a library'', ``In a stadium'',``\underline{In a cinema}''(answer),``At a train station'']
}
\\
\bottomrule
\end{tabular}
}
\caption{Development set examples from the tasks in CLUE. \textbf{Bold} text represents part of the example format for each task. Chinese text is part of the model input, and the corresponding text in \textit{italics} is the English version translated from that. \underline{\textit{Underlined}} text is specially marked in the input. Text in a \texttt{monospaced font} represents the expected model output.}

\label{tab:examples}
\end{table*}
\end{CJK}

\begin{table*}[!t]
\centering
\scalebox{.69}{
\begin{tabular}{l p{1.5cm} p{1.0cm} p{1.5cm} p{1.5cm} p{1.5cm} p{1.5cm} p{1.0cm} p{1.5cm} p{1.5cm} p{1.5cm}}
\toprule
 & Masking & Type & Data Source & Training Tokens \# & Device & Training Steps & Batch Size & Optimizer & Vocabulary & Init Ckpt \\
 \midrule
BERT-base & WordPiece & base & wiki & 0.4B & TPU Pod v2 & - & - & AdamW & 21,128 & Random Init \\

BERT-wwm-ext-base & WWM & base & wiki+ext & 5.4B & TPU v3 & 1M & 384 & LAMB & $\sim$BERT & $\sim$BERT \\

ALBERT-tiny & WWM & tiny & CLUE corpus & 5B & TPU Pod v3 & 500k & 4k & LAMB & $\sim$BERT & Random Init \\

ALBERT-xxlarge & Span & large & CLUE corpus & 5B & TPU Pod v3 & 1M & 8k & AdamW & $\sim$BERT & Random Init \\


ERNIE-base & Knowledge Masking & base & wiki+ext & 15B & NVidia v100 & 1M & 8192 & Adam & 17964 & Random Init \\

XLNet-mid & Sentence Piece & mid & wiki+ext & 5.4B & TPU v3 & 2M & 32 & Adam & 32000 & Random Init \\

RoBERTa-large & WWM & large & CLUE corpus & 5B & TPU Pod & 100k & 8k & AdamW & $\sim$BERT & Random Init \\

RoBERTa-wwm-ext-base & WWM & base & wiki+ext & 5.4B & TPU v3 & 1M & 384 & AdamW & $\sim$BERT & $\sim$BERT \\

RoBERTa-wwm-ext-large & WWM & large & wiki+ext & 5.4B & TPU Pod v3-32 & 2M & 512 & AdamW & $\sim$BERT & Random Init \\

\bottomrule
\end{tabular}
}
\caption{Parameters for pre-training. "BERT-base" is released by google \citep{bert-2019}. ``WWM" stands for whole word masking. ``ext" presents for extended data, different models may use different extended data. ``$\sim$BERT" means similar to Google's Chinese BERT.
\label{tab:app:para:pretrain}}
\end{table*}

\begin{table*}[!t]
\begin{threeparttable}
\scalebox{.9}{
\begin{tabular}{llrrrr}

    \toprule
     & Model & Batch Size & Max Length & Epoch & Learning Rate\\
   
    \midrule

    AFQMC & All* & 16 & 128 & 3 & 2e-5\\
    TNEWS & All* & 16 & 128 & 3 &  2e-5\\

    IFLYTEK & ALBERT-tiny & 32 & 128 & 10 & 2e-5\\
        & RoBERT-large, RoBERTa-wwm-ext-large & 24 & 128 & 3 & 2e-5\\
        & All* except above & 32 & 128 & 3 & 2e-5 \\
   
   OCNLI & BERT-base, RoBERTa-wwm-ext-large & 32 & 128 & 3 & 2e-5 \\
        & RoBERTa-wwm-ext, ERNIE & 32 & 128 & 3 & 3e-5 \\
        & ALBERT-tiny & 32 & 128& 4 &  5e-5\\
        & XLNET-mid & 32 & 128& 3 &  5e-5\\

    
    CLUEWSC2020 & ALBERT-tiny & 8 & 128 & 50 & 1e-4 \\
        & All* except ALBERT-tiny & 8 & 128 & 50 & 2e-5 \\

    CSL 
        & RoBERTa-large & 4 & 256 & 5 & 5e-6 \\
        & All* except above & 4 & 256 & 5 & 1e-5 \\
        
    CMRC* & ALBERT-tiny & 32 & 512 & 3 & 2e-4 \\
          & RoBERTa-wwm-ext-large & 32 & 512 & 2 & 2.5e-5 \\
          & RoBERTa-large & 32 & 256 & 2 & 3e-5 \\
          & XLNET-mid, RoBERTa-wwm-ext-base & 32 & 512 & 2 & 3e-5 \\
          & All* except above & 32 & 512 & 2 & 3e-5 \\
    
    CHID  & All* & 24 & 64 & 3 & 2e-5 \\
    
    C\textsuperscript{3} & All* & 24 & 512 & 8 & 2e-5 \\
    \bottomrule
\end{tabular}
}

\caption{Parameters for fine-tuning. CMRC* presents for CMRC dataset in 2018. All* means ALBERT-tiny, BERT-base, BERT-wwm-ext-base, ERNIE-base, RoBERTa-large, XLNet-mid, RoBERTa-wwm-ext-base and RoBERTa-wwm-ext-large namely. It should be noted that RoBERTa-large is pre-trained with 256 sequence length, which is shorter than 512 length pre-trained for others. So we individually limit the length of RoBERTa-large to 256 for CMRC*, and use the striding text span to relieve this problem. However, this drawback of RoBERTa-large may decrease performances of some datasets whose length can not be effectively cut down, such as C\textsuperscript{3}.}

\label{tab:app:para:fine-tuning}
\end{threeparttable}
\end{table*}


\end{document}